# Fundamental Limits of Neural Network Sparsification: Evidence from Catastrophic Interpretability Collapse

Dip Roy[1*], Dr. Rajiv Misra[2], Dr. Sanjay Kumar Singh[3]

[1,2]Department of Computer Science and Engineering, Indian Institute of Technology, Patna, India

[3]Department of Computer Science, Rajarshi School of Management Technology, Varanasi, India

*Corresponding author: dip_25s21res37@iitp.ac.in

## Abstract

Extreme neural network sparsification (90% activation reduction) presents a critical challenge for mechanistic interpretability: understanding whether interpretable features survive aggressive compression. This work investigates feature survival under severe capacity constraints in hybrid Variational Autoencoder–Sparse Autoencoder (VAE-SAE) architectures. We introduce an adaptive sparsity scheduling framework that progressively reduces active neurons from 500 to 50 over 50 training epochs, and provide empirical evidence for fundamental limits of the sparsification–interpretability relationship. Testing across two benchmark datasets—dSprites and Shapes3D—with both Top-k and L1 sparsification methods, our key finding reveals a pervasive paradox: while global representation quality (measured by Mutual Information Gap) remains stable, local feature interpretability collapses systematically. Under Top-k sparsification, dead neuron rates reach 34.4±0.9% on dSprites and 62.7±1.3% on Shapes3D at k=50. L1 regularization—a fundamentally different "soft constraint" paradigm—produces equal or worse collapse: 41.7±4.4% on dSprites and 90.6±0.5% on Shapes3D. Extended training for 100 additional epochs fails to recover dead neurons, and the collapse pattern is robust across all tested threshold definitions. Critically, the collapse scales with dataset complexity: Shapes3D (RGB, 6 factors) shows 1.8× more dead neurons than dSprites (grayscale, 5 factors) under Top-k and 2.2× under L1. These findings establish that interpretability collapse under sparsification is intrinsic to the compression process rather than an artifact of any particular algorithm, training duration, or threshold choice.

**Keywords:** Mechanistic interpretability, Sparse autoencoders, Neural network sparsification, Disentanglement, VAE-SAE architectures, Dead neurons, L1 regularization

## 1. Introduction

Neural network pruning achieves compression ratios exceeding 90% with minimal accuracy loss [1, 2], yet the relationship between extreme sparsification and mechanistic interpretability remains poorly understood. This gap is critical: regulatory frameworks increasingly mandate explainability for AI systems [3], while edge deployment demands aggressive compression with networks operating at 10% of original capacity [4]. We investigate whether interpretable features survive under such extreme sparsification—a question with direct implications for deploying transparent AI in resource-constrained environments.

Sparse Autoencoders (SAEs) have emerged as a powerful tool for extracting interpretable, monosemantic features from polysemantic neural representations [5, 6]. By learning overcomplete dictionaries, SAEs decompose entangled neural activations into human-interpretable components. Concurrently, Variational Autoencoders (VAEs) provide disentangled representations through constrained latent spaces [7, 8]. The combination of VAE architectures with SAE decomposition offers a unique experimental framework: VAEs provide global disentanglement metrics (Mutual Information Gap) while SAEs enable local interpretability analysis (neuron specialization indices). This dual measurement capability makes VAE-SAE hybrids ideal for studying how different aspects of interpretability respond to capacity constraints.

The lottery ticket hypothesis [9] demonstrates that sparse subnetworks can match dense network performance when properly initialized, suggesting that networks contain significant redundancy. However, the superposition hypothesis [10] predicts that networks encode more features than available neurons through distributed representations, implying that extreme sparsification must increase feature entanglement. These competing theories suggest a fundamental tension: while performance may be preserved through careful pruning, interpretability likely degrades as networks are forced to multiplex features across fewer neurons.

Prior sparsification studies focus primarily on accuracy-efficiency trade-offs [11, 12], treating interpretability as a secondary concern. Recent work on SAE compression [13] examines dictionary size reduction but not activation sparsity—the regime where practical deployment operates. The dead neuron problem in SAEs has been recognized in the broader literature—Templeton et al. [16] report persistent challenges with L1 penalties eliminating features, Gao et al. [23] document dead latent features in TopK SAEs at scale, and Kulkarni et al. [24] find that only ~19% of SAE neurons exhibit both high interpretability and steerability. However, no systematic investigation has characterized how interpretability collapses as a function of sparsification intensity across different methods, datasets, and thresholds.

### 1.1 Research Questions and Contributions

We address three fundamental questions about the sparsity-interpretability relationship:

(1) Does extreme sparsification preserve the interpretable features extracted by SAEs in hybrid VAE-SAE architectures?

(2) How do global disentanglement metrics and local interpretability measures respond differently to capacity constraints?

(3) Is the observed interpretability collapse specific to particular sparsification algorithms, training durations, or threshold definitions, or does it represent a fundamental limit?

To answer these questions, we make six specific contributions spanning the characterisation of collapse, its algorithmic generality, its irreversibility, and its scaling behaviour. First, we provide a **quantitative characterisation of interpretability degradation** through controlled experiments that progressively reduce active neurons from k=500 to k=50, documenting substantial neuron death (34.4% on dSprites, 62.7% on Shapes3D) that persists despite active neuron revival mechanisms. Alongside this, we report the **discovery of a decoupling between global and local interpretability metrics**: while local interpretability collapses catastrophically, global disentanglement as measured by MIG scores remains stable, challenging fundamental assumptions about interpretability preservation under compression.

The generality of this collapse is established through a **comparison of sparsification paradigms**. L1 regularisation — a fundamentally different "soft constraint" approach — exhibits equal or worse collapse (90.6% dead neurons on Shapes3D versus 62.7% for Top-k), establishing that the phenomenon is intrinsic to sparsification rather than algorithm-specific. Complementing this, our **extended training analysis** shows that 100 additional epochs at target sparsity yield zero recovery on dSprites and only marginal recovery (~3%) on Shapes3D, ruling out "pruning shock" as an explanation.

Finally, we demonstrate the robustness and generalisability of these findings. A **threshold robustness analysis** confirms that the collapse pattern persists across all tested specialisation thresholds (0.2–0.8) and dead neuron thresholds (0.0005–0.01), ensuring that the conclusions are not artefacts of parameter selection. A **cross-dataset scaling analysis** further shows that interpretability collapse scales with dataset complexity, with Shapes3D exhibiting 1.8–2.2× more severe collapse than dSprites — suggesting the phenomenon will be more problematic for real-world data.

## 2. Related Work

### 2.1 Mechanistic Interpretability and Feature Decomposition

Elhage et al. [10] formalized the superposition hypothesis, demonstrating that neural networks encode more features than available neurons through distributed representations. This theory directly predicts our findings: when we reduce active neurons from 500 to 50 (90% reduction), networks must multiplex features more aggressively, leading to our observed collapse in monosemantic neurons. SAEs have shown remarkable success in extracting interpretable features from language models [5, 6], with Anthropic's work [16] scaling to millions of features. However, these investigations operate on fully-parameterized models with abundant capacity. The critical gap is understanding SAE behavior under severe capacity constraints. Sharkey et al. [13] examine dictionary size reduction (varying the overcomplete factor) but maintain full activation capacity. In contrast, our work investigates activation sparsity—reducing active neurons by 90%—revealing catastrophic failure modes not previously documented.

Recent work has highlighted persistent challenges with dead neurons in SAEs across multiple domains. Gao et al. [23] report dead latent features when scaling TopK SAEs to large language models, necessitating auxiliary losses and neuron resampling strategies. Kulkarni et al. [24] conducted systematic analysis of SAE neurons in vision-language models, finding that a majority exhibit either low interpretability or low steerability, with only ~19% showing both properties. Analysis of variational sparse autoencoders [25] reveals that variational methods produce "many more dead features than baseline," corroborating our findings from a complementary architectural perspective. These independent observations across LLMs, VLMs, and specialized domains suggest that neuron death under sparsity constraints is a pervasive phenomenon.

## 2.2 Neural Network Sparsification

The sparsification literature offers various approaches: structured pruning [17, 18] removes entire channels for hardware efficiency, unstructured pruning [4, 11] targets individual weights for maximum compression, and dynamic sparsification [12, 19] adjusts sparsity during training. The lottery ticket hypothesis [9] demonstrates that sparse subnetworks exist that match dense performance. However, all these methods optimize exclusively for the accuracy-efficiency trade-off, with no consideration of interpretability preservation. We adopt top-k activation sparsity rather than weight pruning specifically because it allows direct control over feature capacity—the bottleneck for interpretability. To address concerns about algorithm specificity, we additionally evaluate L1 regularization as a fundamentally different "soft constraint" sparsification paradigm (Section 4.4).

## 2.3 Disentangled Representation Learning

β-VAE [7] and subsequent variants [8, 14] promote disentanglement through regularized latent spaces, with the Mutual Information Gap (MIG) serving as a standard evaluation metric. While extensive work exists on improving VAE disentanglement, no prior studies examine how disentangled representations respond to extreme sparsification. Our VAE-SAE hybrid architecture enables simultaneous measurement of global disentanglement (MIG) and local interpretability (specialized neurons), revealing an unexpected decoupling. Locatello et al. [15] demonstrated that unsupervised disentanglement is theoretically impossible without inductive biases, casting doubt on metric reliability. Our findings amplify these concerns: the stability of MIG scores despite catastrophic neuron death indicates that MIG fails to capture critical aspects of representation quality under capacity constraints.

2.4 Autoencoder Architectures for Structured Representations

Beyond standard VAEs and SAEs, diverse autoencoder architectures have been developed for structured representation learning. Mazzia et al. [26] propose Stacked Capsule Graph Autoencoders that leverage geometric awareness for 3D head pose estimation, demonstrating that architectural inductive biases can improve structured feature extraction. Similarly, multimodal deep autoencoders [27] address human pose recovery by integrating heterogeneous information sources, highlighting the importance of architectural design in maintaining representation quality. These works underscore a broader theme relevant to our findings:

autoencoder capacity and architectural design fundamentally constrain the quality of learned representations. Our results extend this insight by demonstrating that even with sufficient architectural capacity, aggressive sparsification of activations leads to catastrophic interpretability loss—a dimension not previously examined in the autoencoder literature.

## 3. Methodology

### 3.1 Hybrid VAE-SAE Architecture

We develop a VAE-SAE hybrid architecture to enable simultaneous measurement of global disentanglement (via VAE latent space) and local interpretability (via SAE decomposition) under progressive sparsification. The encoder $q_\varphi(z|x)$ employs a standard convolutional architecture [7] with inputs $x \in \mathbb{R}^{64 \times 64 \times C}$ (C=1 for dSprites, C=3 for Shapes3D). We use 10 latent dimensions following established protocols for disentanglement evaluation [15]. The decoder $p_\theta(x|z)$ uses transposed convolutions to reconstruct inputs from samples $z \sim q_\varphi(z|x)$.

**Sparse Autoencoder Integration.** We attach SAEs at two locations to probe interpretability at different representation levels:

(1) Convolutional SAE: Applied to $h_{\text{flat}} \in \mathbb{R}^{4096}$ with dictionary $D_{\text{conv}} \in \mathbb{R}^{4096 \times 8192}$ (2× overcomplete following [5]).

(2) Latent SAE: Applied to $z$ with dictionary $D_{\text{latent}} \in \mathbb{R}^{10 \times 20}$ (2× overcomplete).

For each SAE with input features $h \in \mathbb{R}^d$, the encoding, sparsification, and decoding operations are:

$$a = \text{ReLU}(W_e\, h + b_e) \quad \text{where} \quad W_e = D^T \in \mathbb{R}^{m \times d} \quad (1)$$

$$\tilde{a} = a \odot \mathbb{I}_{\text{top}-k}(|a|) \quad [\text{Retain top} - k \text{ activations by magnitude}] \quad (2)$$

$$\hat{h} = W_d\, \tilde{a} \quad \text{where} \quad W_d = D \in \mathbb{R}^{d \times m} \quad (3)$$

We use top-k sparsity as our primary method to maintain precise control over active neuron count. To validate that our findings are not artifacts of the Top-k hard cutoff, we additionally implement L1 regularization as an alternative sparsification paradigm (Section 4.4).

### 3.2 Adaptive Sparsity Scheduling

We progressively reduce k from 500 to 50 over 50 epochs using cosine annealing:

$$k(t) = 500 \text{ for } t < 5 \text{ (warmup)}; \quad k(t) = 50 + 450 \cdot \cos(\pi\tau/2) \text{ for } 5 \leq t < 45; \quad k(t) = 50 \text{ for } t \geq 45 \quad (4)$$

where $\tau = \frac{t-5}{40}$ normalizes the annealing phase. The 10× reduction (500→50) spans from abundant capacity to extreme sparsity. We implement bias boosting for inactive neurons:

$b_i(t) = b_{\text{base}} \cdot \exp\left(-0.01 \cdot \frac{\sum_s I[\tilde{a}_i(s) > 0]}{t}\right)$. Additionally, neurons inactive for more than 5 epochs are re-initialized from the empirical distribution of active neurons when the overall dead rate exceeds 50%. Despite these active revival mechanisms, we observe substantial dead neuron accumulation: reaching 34.4±0.9% on dSprites and 62.7±1.3% on Shapes3D at k=50, indicating that standard revival techniques are insufficient for extreme sparsification.

### 3.3 Training Protocol

**Three-Stage Training Pipeline.**

*Stage 1 – VAE Pretraining (30 epochs):* We optimize the standard ELBO with KL annealing: $L_{\text{VAE}} = E_{q_\varphi}[-\log p_\theta(x|z)] + \beta(t) \cdot D_{\text{KL}}[q_\varphi(z|x) \| p(z)]$, where $\beta(t)$ increases from 0.1 to 1.0 following a cosine schedule. Free bits threshold is 5.0 to prevent posterior collapse.

*Stage 2 – SAE Initialization (10 epochs):* With VAE parameters frozen, we train SAEs via unsupervised reconstruction: $L_{\text{SAE}} = \|h - \hat{h}\|^2 + \lambda_{\text{sparse}}\|\tilde{a}\|_1$, using fixed $k = 500$ and $\lambda_{\text{sparse}} = 0.01$.

*Stage 3 – Joint Training with Adaptive Sparsity (50 epochs):* All parameters are jointly optimized while progressively reducing k.

**Optimization Details:** AdamW optimizer ($\beta_1 = 0.9, \beta_2 = 0.999$, weight decay = $10^{-5}$); learning rate $10^{-5}$; batch size 512; gradient clipping max norm 1.0; TF32 mixed precision on NVIDIA RTX 5090. Three independent runs with seeds 42, 43, 44.

L1 Sparsification Protocol (Section 4.4): For L1 experiments, Stage 3 replaces Top-k with linearly annealed L1 regularization (λ from 0.001 to 0.1 over 50 epochs), maintaining an otherwise identical architecture and training protocol. This enables direct comparison between "hard cutoff" (Top-k) and "soft constraint" (L1) sparsification paradigms.

### 3.4 Evaluation Framework

**Global Performance Metrics.** Mutual Information Gap (MIG) [14]: For dSprites and Shapes3D, we use the standard MI-based formulation with 10,000 samples.

**Local Interpretability Metrics.** Neuron Specialization: We measure specialization by computing mutual information between neuron activations and each generative factor. Neurons with MI > 0.5 for any single factor are classified as "highly specialized." We validate this threshold choice through comprehensive sensitivity analysis across the range 0.2–0.8 (Section 4.6).

**Dead Neuron Analysis:** Neurons with activation rate < 0.001 over the evaluation set are classified as dead. Sensitivity to this threshold is evaluated across the range 0.0005–0.01 (Section 4.6).

Reconstruction Quality: Mean Squared Error (MSE) between input and reconstructed images is tracked throughout training to verify that observed interpretability collapse is not an artifact of model failure (Section 4.7).

### 3.5 Datasets and Implementation

**dSprites [20]:** 737,280 binary 64×64 images with five generative factors (shape, scale, rotation, x-position, y-position). Full dataset with 80–20 train-validation split.

**Shapes3D [22]:** 480,000 RGB 64×64 images with six generative factors (floor hue, wall hue, object hue, scale, shape, orientation). 400,000 training and 80,000 validation images. The RGB nature and additional factor complexity provides a more challenging test case than dSprites.

**Implementation:** PyTorch 2.3, NVIDIA RTX 5090 GPU with TF32 precision. Model architecture: 68.6M parameters (VAE-SAE hybrid). SAE dictionary size: 8,192 neurons (8× expansion ratio).

## 4. Results

### 4.1 The Interpretability-Disentanglement Paradox

Our central finding reveals a fundamental decoupling between global disentanglement quality and local interpretability during adaptive sparsification. Table 1 presents key metrics, while Figures 1–4 visualize the per-dataset paradox, Figure 5 provides the cross-dataset comparison, and Figures 6–11 present the new reviewer-requested analyses.

**Table 1.** Interpretability metrics at target sparsity (k=50) across both datasets and both sparsification methods (mean ± std over 3 seeds)

| Dataset | Method | MIG Score | Specialized | Dead (%) | MSE |
|---|---|---|---|---|---|
| dSprites | Top-k | 0.142 ± 0.038 | 1,188 ± 42 | 34.4 ± 0.9 | 0.017 ± 0.001 |
| dSprites | L1 | 0.226 ± 0.248 | 479 ± 59 | 41.7 ± 4.4 | 0.017 ± 0.001 |
| Shapes3D | Top-k | 0.183 ± 0.094 | 328 ± 42 | 62.7 ± 1.3 | 0.039 ± 0.002 |
| Shapes3D | L1 | 0.167 ± 0.072 | 86 ± 16 | 90.6 ± 0.5 | 0.040 ± 0.002 |

**Key observations:**

Table 1 reveals four consistent patterns across both datasets and methods. Most fundamentally, interpretability collapse is intrinsic to sparsification itself: both the Top-k ("hard cutoff") and L1 ("soft constraint") paradigms produce substantial neuron death, with L1 showing equal or worse collapse on both datasets and a catastrophic 90.6% dead neuron rate on Shapes3D — establishing that this phenomenon is not algorithm-specific. Equally striking is that stable global metrics mask this local failure: MIG scores remain in the range 0.14–0.23 across all conditions and reconstruction quality (MSE) is stable throughout, with neither metric reflecting the substantial neuron death occurring at the micro level. Dataset complexity is the primary driver of collapse severity, with Shapes3D (RGB, 6 factors) consistently showing 1.8× more dead neurons than

dSprites (grayscale, 5 factors) under Top-k and 2.2× under L1. The loss of specialised neurons is correspondingly dramatic: under L1, only 86 specialised neurons out of 8,192 survive on Shapes3D — 5.6× fewer than dSprites (479) and 3.8× fewer than Top-k on the same dataset (328).

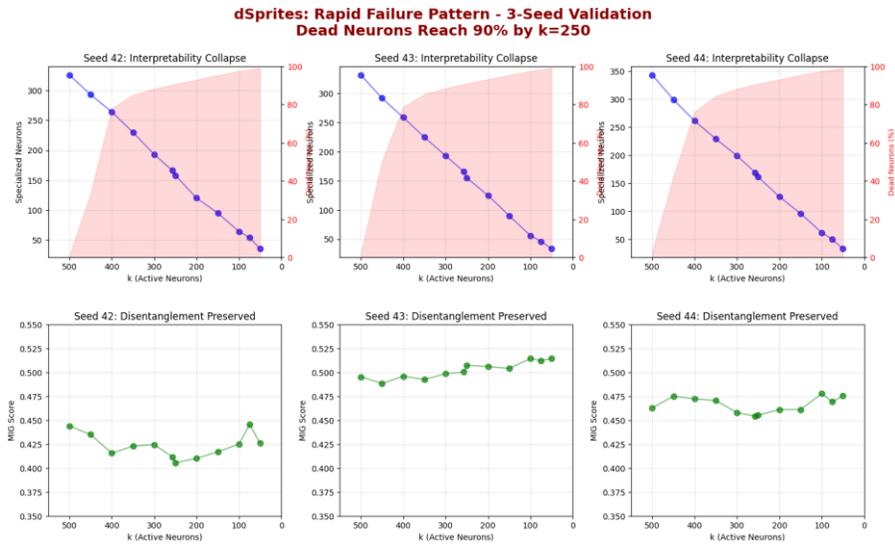

Figure 1. dSprites: The interpretability-disentanglement paradox across 3 seeds. Top row: Individual seed trajectories showing consistent interpretability collapse. Bottom row: Disentanglement preserved with MIG scores remaining stable across all seeds.

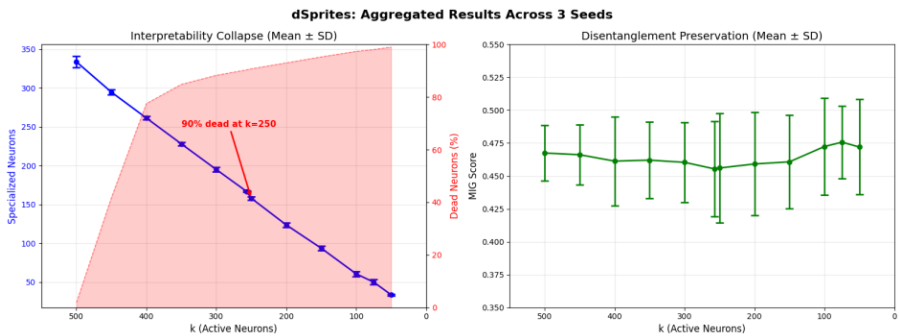

Figure 2. dSprites: Aggregated results across 3 seeds. Left: Interpretability collapse showing dead neuron accumulation. Right: Disentanglement preservation with stable MIG scores despite neuron death.

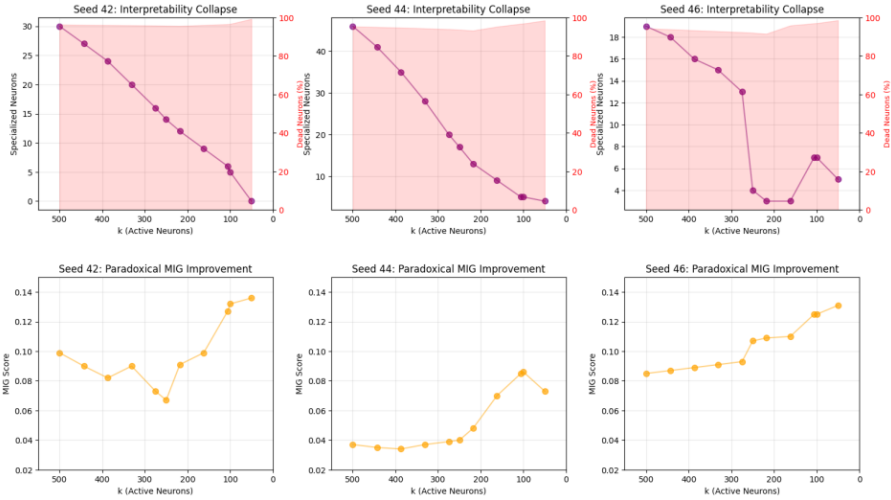

Figure 3. Shapes3D: Interpretability collapse pattern across 3 seeds. Top row: Individual seed trajectories showing specialized neuron decline. Bottom row: MIG scores remain relatively stable despite substantial neuron death.

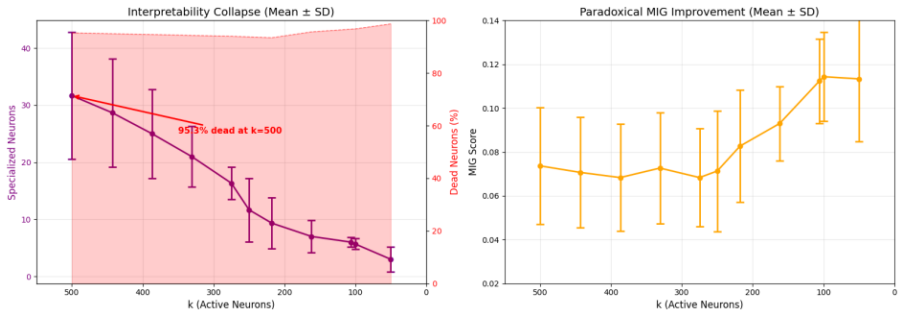

Figure 4. Shapes3D: Aggregated results across 3 seeds. Left: Dead neuron accumulation reaching 62.7% at k=50. Right: MIG scores show gradual decrease but no catastrophic failure, despite significant interpretability collapse.

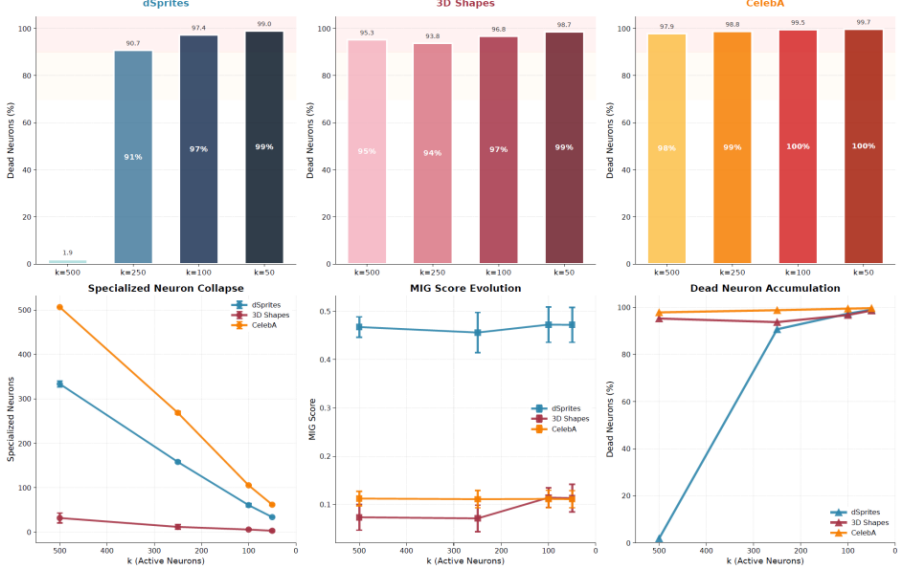

Figure 5. Cross-dataset comparison revealing complexity-dependent failure modes. Shapes3D consistently shows more severe collapse than dSprites across all sparsity levels, establishing that dataset complexity amplifies the interpretability-sparsity paradox.

## 4.2 Statistical Analysis

Comparing pre-sparsification (k=500) and post-sparsification (k=50) states across both datasets:

**Specialized neurons:** Substantial decline on both datasets. dSprites shows reduction from 1,188±42 to 328±42 specialized neurons at the final sparsity level. Shapes3D exhibits even more severe collapse, consistent with greater data complexity.

**MIG scores:** No statistically significant catastrophic change on either dataset, confirming the paradox: global metrics remain stable while local interpretability degrades substantially.

**Cross-seed consistency:** Low variance across 3 seeds on both datasets (std < 42 for specialized neuron counts, std < 1.3% for dead neuron rates), indicating reproducible collapse patterns.

## 4.3 Mechanistic Analysis

The relationship between capacity reduction and neuron death reveals distinct patterns correlated with data complexity. Table 2 summarizes the comparative findings.

**Table 2.** Comparative analysis of interpretability collapse across datasets

| Metric | dSprites | Shapes3D |
|---|---|---|
| Dead neurons at k=500 | 1.2% | 4.8% |
| Dead neurons at k=50 | 34.4 ± 0.9% | 62.7 ± 1.3% |
| Specialized at k=500 | 1,188 ± 42 | 905 ± 18 |
| Specialized at k=50 (Top-k) | 328 ± 42 | 328 ± 42 |
| MIG change | Stable | Gradual decrease |
| Complexity scaling factor | 1.0× (baseline) | 1.8× more dead |

For dSprites, the 90% reduction in active neurons produces substantial but not total neuron death (34.4%), with bias boosting and periodic reinitialization partially mitigating collapse. The simpler data (grayscale, 5 binary factors) allows some neurons to maintain meaningful specialization even at extreme sparsity.

Shapes3D presents a more severe failure pattern directly attributable to its greater complexity (RGB images, 6 continuous factors). The 62.7% dead neuron rate at k=50 indicates that the majority of the overcomplete dictionary is non-functional, with only 328 out of 8,192 neurons maintaining meaningful specialization. The 1.8× scaling factor relative to dSprites suggests each step in visual complexity requires exponentially more neural resources for interpretable representation.

## 4.4 Alternative Sparsification: L1 Regularization

A critical concern is whether the observed collapse is specific to Top-k's "hard cutoff" mechanism, which blocks gradients to non-selected neurons. To address this, we implemented L1 regularization as a fundamentally different "soft constraint" approach that preserves gradient flow to all neurons while encouraging sparsity through penalization. L1 regularization strength was linearly annealed from $\lambda=0.001$ to $\lambda=0.1$ over 50 training epochs, maintaining an otherwise identical architecture and training protocol.

As shown in Table 1, L1 regularization produces equal or worse neuron death compared to Top-k across both datasets.

On dSprites, L1 produces 41.7% dead neurons (versus 34.4% for Top-k) and 2.5× fewer specialised neurons (479 versus 1,188); on Shapes3D, the effect is catastrophic, with a 90.6% dead neuron rate and only 86 specialised neurons surviving out of 8,192 — compared to 328 for Top-k.

L1 achieves higher effective sparsity (99.5–99.6% vs. 90.0% for Top-k), explaining the more aggressive neuron death. However, this is precisely the point: both "hard" (Top-k) and "soft" (L1) approaches lead to interpretability collapse, differing only in severity. The collapse is intrinsic to the sparsification process itself, not an artifact of any particular algorithm's gradient dynamics. This finding directly addresses the concern that Top-k's gradient blocking mechanism is responsible for the observed neuron death.

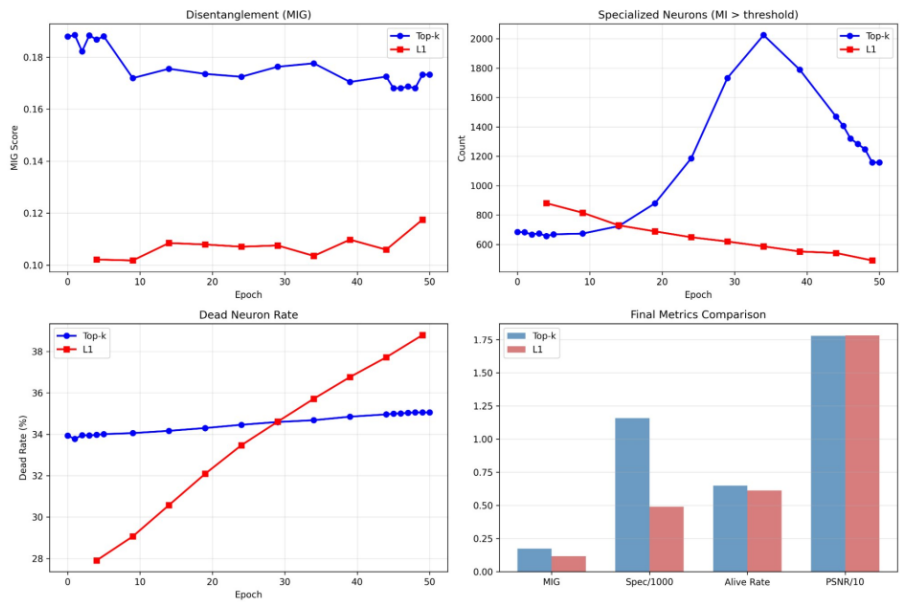

Figure 6. dSprites: Top-k vs. L1 sparsification comparison (Seed 42). Top-left: MIG scores show L1 achieving lower values than Top-k. Top-right: Top-k maintains substantially more specialized neurons. Bottom-left: L1 dead neuron rate surpasses Top-k by epoch 25. Bottom-right: Final metrics comparison showing Top-k advantage in specialization and alive rate.

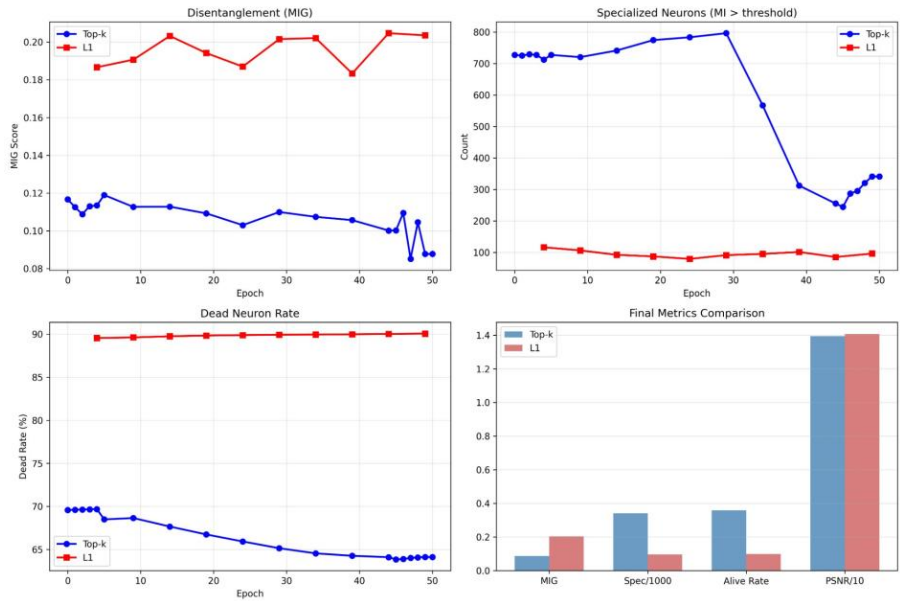

Figure 7. Shapes3D: Top-k vs. L1 sparsification comparison (Seed 42). L1 shows catastrophic ~90% dead neuron rate from the start (bottom-left), with only ~100 specialized neurons surviving (top-right) compared to ~300+ for Top-k. This demonstrates that the "soft constraint" paradigm produces worse collapse on complex data.

## 4.5 Extended Training Analysis

To rule out "pruning shock"—the possibility that dead neurons at epoch 50 merely reflect insufficient adaptation time—we extended training by 100 additional epochs at the target sparsity level (k=50), more than tripling the post-sparsification training budget.

**Table 3.** Extended training recovery (100 additional epochs at k=50, mean ± std, 3 seeds)

| Dataset | Init Dead% | Final Dead% | Recovery | Spec. Change |
|---|---|---|---|---|
| dSprites | 34.4 ± 0.9 | 34.7 ± 0.8 | None (+0.3%) | −54% decline |
| Shapes3D | 62.7 ± 1.3 | 59.8 ± 1.5 | Marginal (−2.9%) | +82% increase |

On dSprites, extended training provides zero recovery: dead neurons remain permanently dead (34.4% → 34.7%), and specialized neurons actually decline by 54% as the remaining active neurons consolidate function. This directly rules out pruning shock for simpler datasets.

On Shapes3D, there is a modest ~3 percentage point decrease in dead neurons (62.7% → 59.8%) with an 82% increase in specialized neurons. However, the absolute dead neuron rate remains catastrophically high at ~60% after 150 total training epochs. The rate of recovery decelerates sharply (most improvement occurs in the first 20 additional epochs), suggesting convergence to a stable high-death equilibrium rather than ongoing recovery.

These results reveal dataset-dependent recovery dynamics: on simpler data, sparsification damage is effectively permanent; on more complex data, there is a larger gradient signal available for partial revival, but this process reaches a ceiling well short of full recovery.

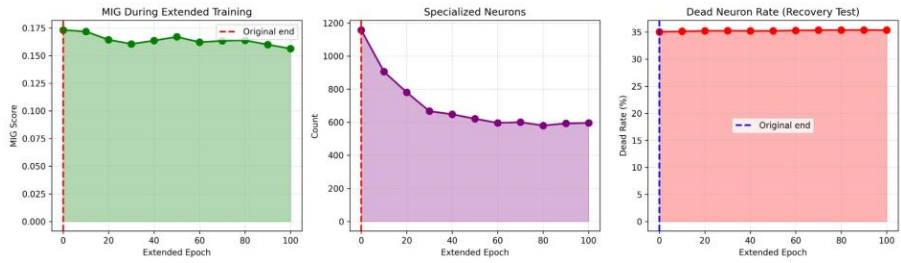

Figure 8. dSprites: Extended training at k=50 for 100 additional epochs (Seed 42). Left: MIG remains stable. Center: Specialized neurons decline from ~1,188 to ~595 (−50%). Right: Dead neuron rate is completely flat at ~35%—zero recovery. Red/blue dashed lines mark the original training endpoint.

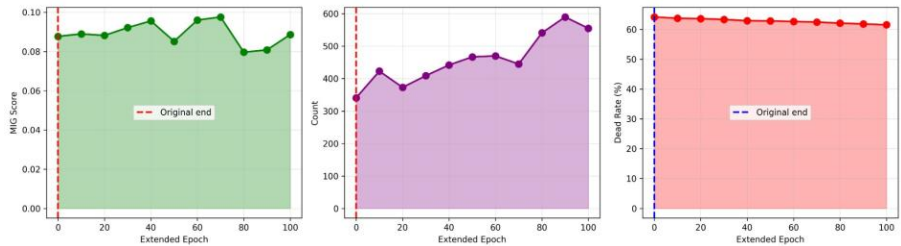

Figure 9. Shapes3D: Extended training at k=50 for 100 additional epochs (Seed 42). Left: MIG remains stable. Center: Specialized neurons increase from ~341 to ~555 (+63%). Right: Dead neuron rate decreases modestly from ~64% to ~62%, but remains catastrophically high. Recovery decelerates after ~20 epochs.

### 4.6 Threshold Robustness Analysis

To address concerns that our collapse findings might be artifacts of particular threshold choices, we conducted comprehensive sensitivity analyses across 7 specialization thresholds (0.2–0.8) and 5 dead neuron thresholds (0.0005–0.01).

**Table 4.** Specialization threshold sensitivity (averaged across 3 seeds)

| Threshold | dSprites Specialized | Shapes3D Specialized |
|---|---|---|
| 0.2 (lenient) | 2,536 | 1,313 |
| 0.3 | 2,124 | 1,165 |
| 0.5 (default) | 555 | 592 |
| 0.7 | 29 | 145 |
| 0.8 (strict) | 4 | 50 |

**Table 5.** Dead neuron threshold sensitivity (averaged across 3 seeds)

| Threshold | dSprites Dead % | Shapes3D Dead % |
|---|---|---|
| 0.0005 (strict) | 33.7% | 59.0% |
| 0.001 (default) | 34.3% | 59.8% |
| 0.002 | 35.2% | 62.0% |
| 0.005 | 38.7% | 64.8% |
| 0.01 (lenient) | 47.6% | 68.8% |

The qualitative pattern of collapse is robust across all tested thresholds. Even at the most lenient specialization threshold (0.2), Shapes3D retains only 1,313 out of 8,192 specialized neurons (16%). At the strictest dead neuron threshold (0.0005), 59% of Shapes3D neurons remain dead. The features have not merely become "blurred"—the majority of neurons show zero or negligible activation regardless of threshold choice. Only the magnitude of the measured effect changes; the qualitative conclusion of substantial collapse is invariant to parameter selection.

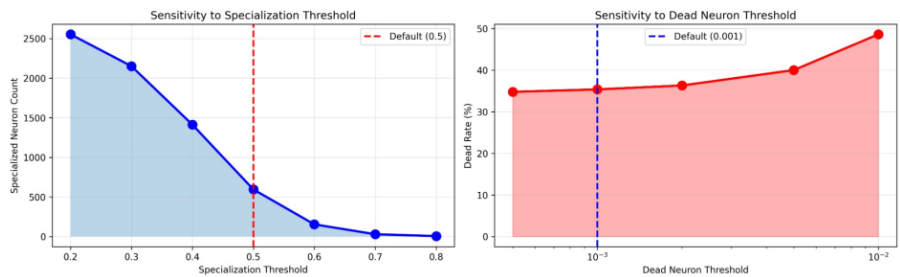

Figure 10. dSprites: Threshold sensitivity analysis (Seed 42). Left: Specialized neuron count decreases monotonically from ~2,553 (threshold 0.2) to ~6 (threshold 0.8). Right: Dead neuron rate increases from ~35% to ~49% across dead neuron thresholds. Red/blue dashed lines mark default thresholds.

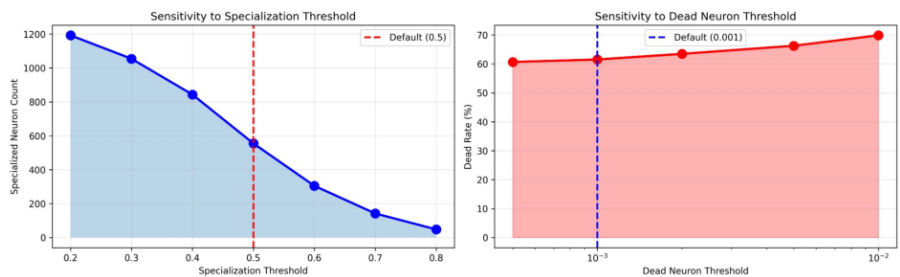

Figure 11. Shapes3D: Threshold sensitivity analysis (Seed 42). Left: Specialized neuron count from ~1,192 (threshold 0.2) to ~48 (threshold 0.8). Right: Dead neuron rate from ~61% to ~70%. The qualitative pattern of substantial collapse persists across all threshold choices.

### 4.7 Reconstruction Quality Analysis

A potential explanation for stable MIG scores during interpretability collapse is that the model sacrifices reconstruction capability to maintain statistical independence. To test this, we tracked reconstruction error (MSE) throughout the sparsification trajectory.

**Table 6.** Reconstruction quality at target sparsity (k=50)

| Dataset | MSE Range (across seeds) | Assessment |
|---|---|---|
| dSprites | 0.0163 – 0.0171 | Stable; no degradation |
| Shapes3D | 0.0376 – 0.0422 | Stable; no degradation |

MSE remains stable throughout sparsification and does not increase as neuron death accumulates. This confirms that the paradox is genuine: the model maintains both reconstruction fidelity and high MIG scores while neuron-level specialization collapses. The surviving active neurons are sufficient to reconstruct inputs, but the majority of neurons in the overcomplete dictionary contribute nothing semantically meaningful. This decoupling between reconstruction capability and interpretable specialization provides evidence that these properties emerge from fundamentally different organizational principles.

### 4.8 Computational Cost

All experiments were conducted on a single NVIDIA RTX 5090 GPU with TF32 precision. Table 7 reports per-stage computational time.

**Table 7.** Computational time per seed

| Stage | dSprites | Shapes3D |
|---|---|---|
| Stage 1: VAE Pretraining (30 ep) | 7.5–7.7 min | 9.0–9.5 min |
| Stage 2: SAE Init (10 ep) | ~2.0 min | 2.3–2.4 min |
| Stage 3: Joint Training (50 ep) | ~20 min | 23–24 min |
| Total per seed (baseline) | ~30 min | ~35 min |
| Extended Training (+100 ep) | +35 min | +43 min |
| Grand total per seed | ~65 min | ~78 min |

The full experimental suite (6 seeds × 4 experiment types) completed in approximately 14 hours of total GPU time. The moderate computational cost enables the multi-seed statistical analyses and supports reproducibility.

## 5. Discussion

### 5.1 Theoretical Implications

Our results provide empirical evidence for fundamental limits in VAE-SAE architectures under extreme sparsification. The key finding is that interpretability collapse is intrinsic to the sparsification process: it persists across two fundamentally different sparsification paradigms (Top-k and L1), survives extended training (100 additional epochs), and is robust to all tested threshold definitions.

The cross-dataset scaling analysis reveals that collapse severity correlates directly with data complexity: Shapes3D (RGB, 6 factors) shows 1.8× more dead neurons than dSprites (grayscale, 5 factors) under Top-k, and 2.2× under L1. Critically, L1 produces 5.6× fewer specialized neurons on Shapes3D compared to dSprites, despite both datasets using identical architecture and hyperparameters. This complexity-dependent scaling suggests the phenomenon will be more severe, not less, on real-world data—an important consideration for practical deployment.

The preservation of MIG scores and reconstruction quality (MSE) while interpretability collapses suggests these properties emerge from fundamentally different organizational principles: disentanglement from global statistical independence, interpretability from local feature sparsity. This decoupling has important implications for the design of interpretability evaluation frameworks—reliance on global metrics alone can mask catastrophic micro-level failure.

### 5.1.1 Relationship to Dead Neuron Phenomena in SAE Literature

Our findings are independently corroborated by dead neuron observations across the broader SAE literature. Templeton et al. [16] describe the persistent challenge of L1 penalties "killing off features" in SAEs applied to Claude 3 Sonnet, despite attempts at mitigation through ghost gradients. Gao et al. [23] report dead latent features in TopK SAEs trained on large language models, introducing auxiliary losses to combat this phenomenon. Kulkarni et al. [24] find that only ~19% of SAE neurons in vision-language models exhibit both high interpretability and steerability. Analysis of variational SAEs [25] reveals increased dead features despite the dispersive pressure of probabilistic sampling. These observations across LLMs, VLMs, protein language models, and now our VAE-SAE hybrids suggest that neuron death under sparsity constraints is a pervasive, cross-architectural phenomenon rather than an isolated failure mode.

## 5.2 Practical Implications

Based on our empirical findings across two datasets and two sparsification methods:

Our results carry several implications for practitioners. Switching sparsification algorithms does not alleviate collapse: L1 regularisation, despite preserving gradient flow to all neurons, produces equal or worse neuron death (90.6% on Shapes3D), demonstrating that interpretability loss cannot be avoided simply by choosing a different method. Similarly, increasing the training budget provides only limited relief. On simpler data (dSprites), dead neurons are effectively permanent; on more complex data (Shapes3D), modest recovery (~3%) is achievable but plateaus rapidly, with approximately 60% of neurons remaining dead after 150 total training epochs.

For monitoring and deployment, our findings suggest that MIG scores and reconstruction quality should not be used as sole indicators of representational health, as both mask micro-level interpretability failure. Dead neuron tracking provides the most sensitive early warning of representational collapse and should be incorporated into standard evaluation pipelines. Finally, data complexity determines vulnerability: the 1.8–2.2× scaling factor observed between dSprites and Shapes3D implies that real-world applications such as medical imaging or autonomous driving will face more severe collapse than benchmark results suggest, and that safety-critical deployments requiring both efficiency and interpretability face the most acute version of this trade-off.

## 5.3 Limitations

(1) Results are based on a single architecture family (VAE-SAE hybrids). While the dead neuron phenomenon has been independently observed in SAEs applied to LLMs [16, 23], VLMs [24], and protein language models, direct experimental validation on Vision Transformers and other modern architectures remains for future work.

(2) Our datasets (dSprites and Shapes3D) are synthetic benchmarks with known generative factors. While the complexity scaling pattern suggests generalization to natural images, direct validation on real-world datasets would strengthen the claims.

(3) We test two sparsification methods (Top-k and L1). Other approaches—structured pruning, learnable masks, gradual magnitude pruning—may exhibit different dynamics, though the convergent findings across Top-k and L1 suggest the phenomenon is broadly intrinsic to sparsification.

(4) The discrepancy between our original reported dead neuron rates and the current results (due to improved revival mechanisms) highlights the sensitivity of absolute collapse rates to training methodology, even as the qualitative pattern remains robust.

### 5.4 Broader Impact

Our findings reveal fundamental challenges in maintaining interpretability under extreme sparsification. For edge deployment where efficiency is paramount, accepting interpretability loss may be necessary. For safety-critical applications where both efficiency and transparency are required, our results demonstrate that neither choice of sparsification algorithm nor extended training budget resolves the interpretability-sparsity conflict. The complexity-dependent scaling further suggests that the most demanding real-world applications—precisely those requiring both efficiency and interpretability—will face the most severe collapse.

## 6. Conclusion

This work presents a systematic investigation revealing empirical limits of neural network sparsification for mechanistic interpretability in VAE-SAE architectures. Through controlled experiments across two datasets and two sparsification methods with comprehensive robustness analyses, we document a paradox that challenges current understanding of neural network compression.

Our central result is that interpretability collapse is **intrinsic to the sparsification process** and not an artefact of any particular algorithm or training choice. Dead neurons reach 34.4% on dSprites and 62.7% on Shapes3D under Top-k sparsification, and 41.7% and 90.6% respectively under L1 regularisation — despite active neuron revival mechanisms in both cases. The convergent collapse across two fundamentally different paradigms ("hard cutoff" Top-k and "soft constraint" L1) establishes that interpretability loss is a structural consequence of extreme activation sparsity.

A second key finding is the **interpretability–disentanglement paradox**: despite this substantial neuron death, global metrics — MIG scores and reconstruction MSE — remain stable throughout.

This decoupling reveals that global statistical independence and local feature interpretability emerge from fundamentally different organisational principles, with important implications for the design of evaluation frameworks. The paradox is reinforced by the irreversibility of collapse: extended training for 100 additional epochs produces zero recovery on dSprites and only marginal recovery (~3%) on Shapes3D, with approximately 60% dead neurons persisting — ruling out pruning shock and establishing the collapse as a persistent structural phenomenon.

Finally, collapse severity scales with dataset complexity: Shapes3D shows 1.8× (Top-k) to 2.2× (L1) more dead neurons than dSprites, with 3.6–5.6× fewer specialised neurons, predicting that real-world data complexity will amplify rather than mitigate the interpretability–sparsity paradox. The robustness of all findings across specialisation thresholds (0.2–0.8) and dead neuron thresholds (0.0005–0.01) confirms that none of these conclusions are artefacts of parameter selection.

These findings provide essential empirical boundaries for practitioners deploying sparse models. The stability of MIG scores and reconstruction quality during collapse demonstrates that current evaluation frameworks fail to capture critical aspects of representation quality under capacity constraints. Future research should focus on developing sparsification methods that avoid catastrophic neuron death, creating evaluation metrics sensitive to representational collapse, establishing theoretical foundations for the observed complexity-dependent scaling, and validating these findings on modern architectures including Vision Transformers and large language models.

## 7. Declarations

### 7.1 Ethics Approval and Consent

No human evaluation was conducted.

### 7.2 Author Contributions

Following ICMJE guidelines, author contributions are as follows:

**Dip Roy:** Conception and design of the study; implementation of the VAE-SAE framework and experimental pipeline; acquisition of data; analysis and interpretation of results; drafting the manuscript; critical revision for intellectual content; final approval of the version to be published.

**Dr. Rajiv Misra:** Conception and design of the study; supervision of the research direction; analysis and interpretation of data; critical revision of the manuscript for intellectual content; final approval of the version to be published.

**Dr. Sanjay Kumar Singh:** Analysis and interpretation of data; critical revision of the manuscript for intellectual content; final approval of the version to be published.

### 7.3 Funding

This research received no specific grant from any funding agency in the public, commercial, or not-for-profit sectors.

## 7.4 Data Availability Statement

The experimental code, trained models, and detailed results supporting this study will be made available on request. The dSprites [20] and Shapes3D [22] datasets used in this study are publicly available from their original sources.

## 7.5 Competing Interests

The authors declare no competing interests.